\documentclass[10pt,twocolumn,letterpaper]{article}

\usepackage{wacv}
\usepackage{times}
\usepackage{epsfig}
\usepackage{graphicx}
\usepackage{amsmath}
\usepackage{amssymb}

\usepackage[sort,numbers]{natbib}

\usepackage{multirow}
\usepackage[table]{xcolor}
\usepackage{algorithm2e}
\usepackage{bbm}
\usepackage{xparse}

\newcommand{\ircnn}{IRCNN~\cite{ircnn}}
\newcommand{\dncnn}{CDnCNN~\cite{dncnn}}
\newcommand{\ffdnet}{FFDNet~\cite{ffdnet}}
\newcommand{\bmd}{*CBM3D~\cite{bm3d}}

\usepackage[pagebackref=true,breaklinks=true,letterpaper=true,colorlinks,bookmarks=false]{hyperref}

\wacvfinalcopy

\begin{document}

\title{Identifying Recurring Patterns with Deep\\%
 Neural Networks for Natural Image Denoising}

\author{Zhihao Xia, Ayan Chakrabarti\\Washington University in St. Louis\\{\tt \small \{zhihao.xia,ayan\}@wustl.edu}}

\maketitle
\begin{abstract}
  Image denoising methods must effectively model, implicitly or explicitly, the vast diversity of patterns and textures that occur in natural images. This is challenging, even for modern methods that leverage deep neural networks trained to regress to clean images from noisy inputs. One recourse is to rely on ``internal'' image statistics, by searching for similar patterns within the input image itself.  In this work, we propose a new method for natural image denoising that trains a deep neural network to determine whether patches in a noisy image input share common underlying patterns. Given a pair of noisy patches, our network predicts  whether different sub-band coefficients of the original noise-free patches are similar. The denoising algorithm then aggregates matched coefficients to obtain an initial estimate of the clean image. Finally, this estimate is provided as input, along with the original noisy image, to a standard regression-based denoising network. Experiments show that our method achieves state-of-the-art color image denoising performance, including with a blind version that trains a common model for a range of noise levels, and does not require knowledge of level of noise in an input image. Our approach also has a distinct advantage when training with limited amounts of training data.
\end{abstract}
\section{Introduction}
\label{sec:intro}

The sheer diversity of content, that can be present in photographs of natural scenes, makes them a challenge for algorithms that must model their statistics for various image restoration tasks, including the classical task of image denoising: recovering an estimate of a clean image from a noisy observation. A common approach is to rely on image models for local image regions---either explicitly as parametric priors or implicitly as estimators trained via regression---with parameters learned on databases of natural images~\cite{svd,foe,foe0,epll,mlp,tnrd,ircnn,dncnn,ffdnet,dboost}.

An important class of methods adopt a different modeling approach, to exploit self-similarity in images by relying on their ``internal statistics''~\cite{nlmeans,bm3d0,bm3d}. A particularly successful example from this class is the BM3D~\cite{bm3d0,bm3d} algorithm, which identifies sets of similar patches in noisy images using sum of squared distances (SSD) as the matching metric, and then uses the statistics of each set to denoise patches in that set. Applying this process twice, BM3D produced high-quality estimates that, until recently,  represented the state-of-the-art in image denoising performance.

However, recent methods have been able to exceed this performance by using neural networks trained to regress to clean image patches from noisy ones~\cite{mlp,tnrd,ircnn}. With carefully chosen architectures, these methods are able to use the powerful expressive capacity of neural networks to better learn and  encode image statistics from external databases, and thus exceed the capability of self-similarity based methods. In this work, we describe a denoising method that brings the expressive capacity of neural networks to the task of identifying and leveraging recurring patterns in the underlying images from their noisy observations.

Our contributions are as follows:\vspace{-0.5em}
\begin{itemize}
\item We introduce a novel matching network that looks at pairs of noisy patches at a time, and makes fine-grained predictions of the similarity of their underlying clean versions. Specifically, our network outputs \emph{separate} matching scores for different groups of wavelet coefficients, to exploit similarities that exist at some orientations and scales, but not others. These scores are used for averaging to form an initial denoised estimate.\vspace{-0.5em}
\item We propose a \emph{two-step process} to train this matching network, with respect to denoising quality, that leads to convergence to a better network model.\vspace{-0.5em}
\item We combine the matching network with a standard regression step to yield a complete algorithm that achieves \emph{state-of-the-art} denoising performance.\vspace{-0.5em}
\item We carry out extensive experiments on multiple datasets and show that our method consistently yields higher quality estimates than the state-of-the-art on a variety of metrics. Indeed, even a \emph{blind} version of our model---that does not have knowledge of the noise level---outperforms state-of-the-art methods that do.\vspace{-0.5em}
\item Finally, we show that our method has a distinct advantage over regression-based networks when learning from only a small amount of training data. In these cases, our method is able to generalize better due to its reliance on per-image internal statistics.
\end{itemize}
\begin{figure*}[!t]
  \centering
  \includegraphics[width=\textwidth]{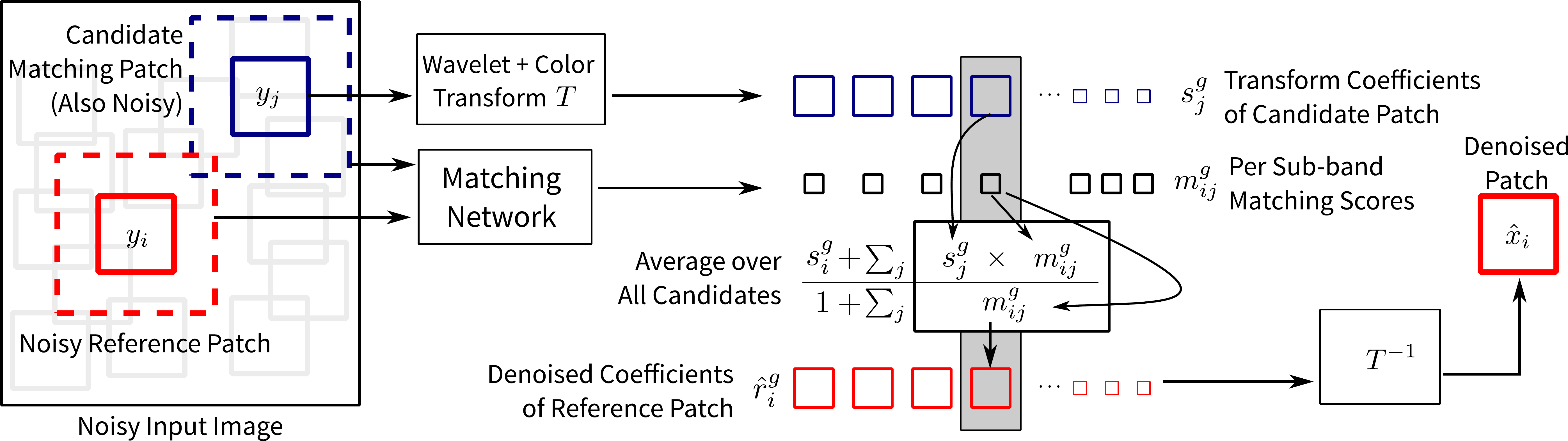}
  \caption{Overview of Our Approach to Patch Denoising. We produce estimates of clean patches by weighted averaging across a candidate set of nearby patches in the observed noisy input. We decompose every patch using a wavelet and de-correlating color transform into sets of sub-band coefficients, with coefficients at the same scale, orientation, and color channel grouped together in each set. We then train a neural network that, given a pair of patches, computes a vector of matching scores---one for each group of coefficients. For every patch, we compute these score vectors with respect to all candidate patches. The denoised patch is obtained by averaging, across all candidates, of each group of coefficients using its corresponding matching scores.}
  \label{fig:teaser}
\end{figure*}

\section{Related Work}
\label{sec:rw}

Denoising is a classical problem in image restoration. In addition to its practical utility in improving the quality of photographs taken in low-light or by cheaper sensors,  image denoisers can be used as generic ``plug-and-play'' priors within iterative approaches to solve a larger variety of generic image restoration tasks (e.g.,~\cite{plugplay,onenet,ircnn,meanshift}).

Many classical approaches to image denoising are based on exploiting statistics of general natural images, using estimators or explicit statistical priors~\cite{svd,foe,foe0,epll}, whose parameters are learned from datasets of clean images. A different category of approaches use patch-recurrence of self-similarity~\cite{nlmeans,bm3d0}, to address the fact that there is significant diversity in content across images, while the variation within a \emph{specific} image is far more limited. There is a natural trade-off between these two approaches: methods based on external statistics have the opportunity to learn them from clean images, but these statistics may be too general for a specific image; while those based on self-similarity work with a more relevant model for each image, but must find a way to derive its parameters from the noisy observation itself. We refer the reader to the work of Mosseri et al.~\cite{irani} for an insightful discussion.

BM3D~\cite{bm3d0} is a particularly successful method that is based on self-similarity. It works by organizing similar patches into groups (using SSD as the matching metric, and doing two rounds of matching), and denoising patches based on the statistics of its group through collaborative filtering. However, recent algorithms~\cite{mlp,tnrd,ircnn,ffdnet,dncnn,dboost} have been able to surpass BM3D's performance using estimators trained on external datasets, leveraging the powerful implicit modeling capacity of deep neural networks.

In this work, we propose a new method that uses neural networks to identify and exploit self-similarity in noisy images. This was also the goal of methods by Lefkimmiatis~\cite{nlcnn} and Yang and Sun~\cite{bm3dnet}, who proposed interesting approaches that are based on designing network architectures that ``un-roll'' and carry out the computations in BM3D and non-local means denoising, and then train the parameters of these steps discriminatively through back-propagation, resulting in performance gains over the baseline methods. More recently, \cite{liu2018non} and \cite{plotz2018neural} proposed using non-local aggregation steps on the intermediate activations of their network. In contrast, we employ a substantially different approach: denoising in our framework is achieved by weighted averaging of different sub-band coefficients of the noisy image patches themselves, while our network is tasked with predicting optimal values of these weights by matching. Also note that, unlike \cite{plotz2018neural} which makes a hard selection of a small number of the best matches for every reference location, we average across a dense set of matches in a large neighborhood, with different continuous weights.

The primary component of our method is a network that must learn to match patches through noise. Several neural network-based methods have been proposed to solve matching problems~\cite{lecun,str1,str2}, with the goal of finding correspondences across images for tasks like stereo. Our method is motivated by their success, and we borrow several design principles of their architectures. However, our matching network has a completely different task: denoising. Our network is thus trained with a loss optimized for denoising (as opposed to classification or triplet losses), and instead of predicting a single matching score for a pair of patches, produces a richer description of their commonality  with distinct scores for different sub-bands.

\newcommand{\tilx}{r}
\newcommand{\tily}{s}

\section{Proposed Denoising Algorithm}
\label{sec:method}

Our goal is to produce an estimate of an image $X$ given observation $Y$ that is  degraded by i.i.d.~Gaussian noise, i.e.,
\begin{equation}
  \label{eq:obsmodel}
Y=X+\epsilon, \epsilon\sim\mathcal{N}(0,\sigma_z^2 I).  
\end{equation}
Our algorithm leverages the notion that many patterns will occur repeatedly in different regions in the underlying clean image $X$, while the noise in those regions in $Y$ will be un-correlated and can be attenuated by averaging. In this section, we describe our approach to training and using a deep neural network to identify these recurring patterns from the noisy image, and forming an initial estimate of $X$ by averaging matched patterns. We then use a second network to regress to the final denoised output from a combination of these initial estimates and the original noisy observation.

\subsection{Denoising by Averaging Recurring Patterns}

Our initial estimate is formed by denoising individual patches in the image, by computing a weighted average over neighboring noisy patches with weights provided by a matching network. Formally, given the noisy observation $Y$ of an image $X$, we consider sets of overlapping patches $\{y_i = P_iY\}$ (corresponding to clean versions $\{x_i = P_iX\}$), where each $P_i$ is a linear operator that crops out intensities of a different square patch  (of size $8 \times 8$ in our implementation) from the image. We then use a de-correlating color space followed by a Harr wavelet transform to obtain coefficients $\tily_i = Ty_i$ (corresponding to clean versions $\tilx_i=Tx_i$), where the orthonormal matrix $T$ represents the color and wavelet transforms.

We group these coefficients into sets $\{\tily_i^g\}$ where each set includes all coefficients with the same orientation (horizontal, vertical, or diagonal derivative), scale or pyramid level, and color channel\footnote{For $8\times 8$ patches, this gives us 30 coefficient groups: 27 corresponding to 3 color channels, 3 scales, and 3 derivative orientations; and an additional 3 coefficients for the scaling coefficients of the 3 color channels.}%
. Then, for every patch $y_i$ we consider a set of candidate matches composed of other noisy patches in the image $y_j, j \in \mathcal{N}_i$, from a large neighborhood around $i$. As illustrated in Fig.~\ref{fig:teaser}, our method produce an estimate of the denoised coefficients $\hat{\tilx}_i$ as a weighted average of the corresponding coefficients of the candidate patches:
\begin{equation}
  \label{eq:wavg}
  \hat{\tilx}_i^g = \left(1+ \sum_{j \in \mathcal{N}_i} m_{ij}^g\right)^{-1}\left(\tily_i^g + \sum_{j \in \mathcal{N}_i} m_{ij}^g \tily_j^g\right),
\end{equation}
where $m^g_{ij} \geq 0$ are scalar matching weights that are a prediction of the similarity between the $g^{th}$ set of coefficients in patches $i$ and $j$ respectively.

This gives us a denoised estimate for each patch $\hat{x}_i$ in the image as $T^{-1}\hat{\tilx}_i$. We then obtain an estimate $\hat{X}$ of the full clean image simply by averaging the denoised patches $\hat{x}_i$, i.e., the denoised estimate of each pixel is computed as the average of its estimate from all patches that contain it.

\subsection{Predicting Matches from Noisy Observations}

The success of our match-averaging strategy in \eqref{eq:wavg} depends on obtaining optimal values for the matching scores $m_{ij}^g$.  Intuitively, we want $m_{ij}^g$ to be high when the \emph{clean} coefficients $\tilx_i^g$ and $\tilx_j^g$ are close, so that the averaging in \eqref{eq:wavg} will attenuate noise and yield $\hat{\tilx}_i^g$ close to $\tilx_i^g$. Conversely, we want $m_{ij}^g$ to be low where the two sets of underlying clean coefficients are not similar, because averaging them would yield poor results, potentially worse than the noisy observation itself. However, note that while the optimal values of these matching scores depend on the characteristics of the clean coefficients $\{\tilx_i^g\}$, we only have access to their noisy counterparts $\{\tily_j^g\}$.

Therefore, we train a neural network $\mathcal{M}$ to predict the matching scores given a pair of larger noisy patches ($16\times 16$ in our implementation) $y_i^+$ and $y_j^+$ centered around $y_i$ and $y_j$ respectively: $m_{ij} = \mathcal{M}(y_i^+,y_j^+)$, where $m_{ij} = [\ldots,m_{ij}^g,\ldots]$ is a vector of matching scores for all sets of coefficients. We don't require the output of the network $\mathcal{M}$ to be symmetric ($m_{ij}$ need not be the same as $m_{ji}$), and we use the same network model for evaluating all patch pairs, being agnostic to their absolute or relative locations.

\begin{figure*}[!t]
  \centering
  \includegraphics[width=\textwidth]{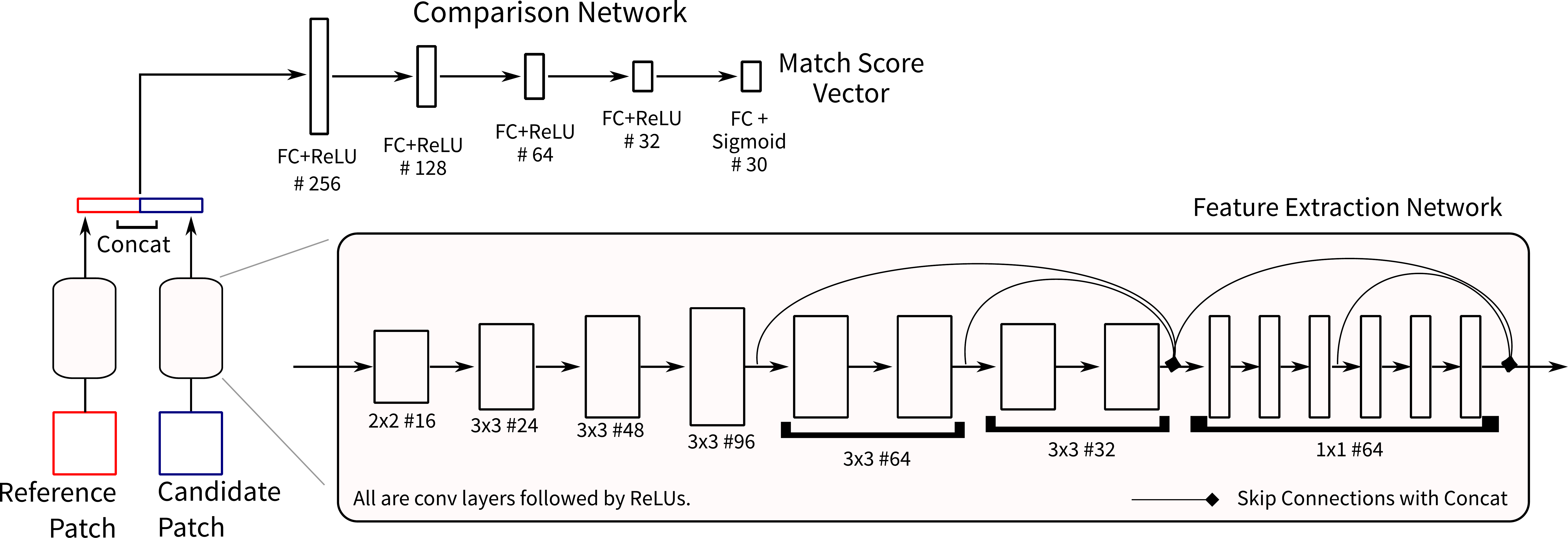}
  \caption{Matching Network Architecture. To produce the matching scores $m_{ij}^g$, we first extract a feature vector for all patches by passing the image through a feature extraction network, comprised of a set of convolutional layers with skip connections (where the join is performed by a concatenate operation. Then, for any pair of patches, we take the corresponding pair of feature vectors, and pass them after concatenation through a series of fully-connected layers. The final layer has a sigmoid activation, yielding scores that lie between 0 and 1.}
  \label{fig:arch}
\end{figure*}

The matching network $\mathcal{M}$ has a Siamese-like architecture as illustrated in Fig.~\ref{fig:arch}. It begins with a common feature extraction sub-network applied to both input patches to produce a feature-vector for each. This sub-network has a receptive field of $16\times 16$, and includes a total of fourteen convolutional layers with skip connections~\cite{densenet} including at the final output (see Fig.~\ref{fig:arch}). The computed feature-vectors for each of the two inputs are then concatenated and passed through a comparison sub-network, which comprises of a set of five fully-connected layers. All layers have ReLU activations, except for the last which uses a sigmoid to produce the match-scores $m_{ij}^g$. These scores are thus constrained to lie in $[0,1]$. Note that during inference, the feature extraction sub-network needs to be applied only once to compute feature-vectors for all patches in a fully-convolutional way. Only the final five fully connected layers need to be repeatedly applied for different patch pairs.

Observe that our matching network takes the noisy patches directly as input, while using the wavelet and color transforms to parameterize its outputs, as a means of providing a more fine-grained characterization of similarity between patches than a single score. Moreover, although the inputs to the network itself consist only of a pair of relatively small patches, it enables aggregation from a dense set of patches in a large neighborhood, through repeated application on multiple patch pairs and averaging as per \eqref{eq:wavg}.

\subsection{Training}
We train the matching network $\mathcal{M}$  to produce matching scores that are optimal with respect to the quality of the denoised patches $\hat{x}_i$. Specifically, we use an $L_2$ loss between the true and estimated clean patches:
\begin{equation}
  \label{eq:loss}
L=\|x_i-\hat{x}_i\|^2=\sum_g \|\hat{\tilx}_i^g -{\tilx}_i^g\|^2,
\end{equation}
where the denoised coefficients $\hat{\tilx}_i^g$ are computed using \eqref{eq:wavg} based on matching-scores predicted by the network. Note that the loss for a single patch $x_i$ will depend on matching scores produced by the network for $x_i$ paired with all candidate patches in its neighborhood $\mathcal{N}_i$.

While it is desirable to train the network in this end-to-end manner with our actual denoising approach, we empirically find that training the network with this loss from a random initialization often converges to a sub-optimal local minima. We hypothesize that this is because we compute gradients corresponding to a large number of matching scores (all candidates in $\mathcal{N}_i$) with respect to supervision only from a single denoised patch.

Thus, we adopt a pre-training strategy using a loss defined on pairs of patches at a time, using a simplified loss for denoising patch $i$ by averaging it with patch $j$ as:
\begin{equation}
  \label{eq:modloss}
  \hat{L}_{ij} = \sum_g \frac{\|\tily_i^g-\tilx_i^g\|^2 + {m_{ij}^g}^2\|\tily_j^g-\tilx_i^g\|^2}{\left(1+{m_{ij}^g}\right)^2}.
\end{equation}
This is equivalent to the actual loss in \eqref{eq:loss} with performing the averaging in \eqref{eq:wavg} with only one candidate patch $j$, by dropping the cross term between $(x_i-y_i)$ and $(x_i-y_j)$, i.e., by assuming the noise is un-correlated with the difference between the two patches. It is interesting to note here if we assume that the deviations between reference and candidate patches are un-correlated, for different candidates $j \in \mathcal{N}_i$, then the optimal averaging weight for a given candidate is the same whether averaging with one or multiple candidates. The modified loss in \eqref{eq:modloss} serves as a good initial proxy for pre-training, but since the un-correlated deviation assumption does not hold in practice, we follow this with training with the actual loss in \eqref{eq:loss}.

In particular, we pre-train the network for a number of iterations using the modified loss in \eqref{eq:modloss}---constructing the training set by considering all non-overlapping patches $i$ in an image, with random shuffling to select candidate $j$ for each patch $i$, and train with respect to the loss of both matching $i$ to $j$ and vice-versa. This allows us to compute updates with respect to a much more diverse set of patches into a training batch, and to make maximal use of the feature extraction computation during training. The pre-training step is followed by training the network with the true loss in \eqref{eq:loss} till convergence---here, we extract a smaller number of training reference patches from each image, along with \emph{all} their neighboring candidates.

\subsection{Final Estimates via Regression}

While the initial estimates produced by our method as described above are of reasonable quality, they are limited by \eqref{eq:wavg} restricting every denoised patch to be a weighted average of observed noisy patches. To overcome this and achieve further improvements in quality, we use a second network trained via traditional regression to derive our final denoised estimate. Specifically, we adopt the architecture of IRCNN~\cite{ircnn} with has seven dilated convolutional layers. In our case, this network takes a six-channel input formed by concatenating the original noisy input and our initial denoised estimate from match-based averaging. The output of the last layer is interpreted as a residual, and added to the initial estimates to yield the final denoised image output.

After the matching network has been trained, we generate sets of clean, noisy, and initial denoised estimates. This serves as the training set for this second network which is trained using an $L_2$ regression loss. We find that this step leads to further improvement over our initial estimates, while also outperforming state-of-the-art denoising networks (including IRCNN~\cite{ircnn} itself).

\newcommand{\pars}[1]{\vspace{-1em}\paragraph{#1}}

\begin{table*}[!t]\small
  \centering
  \setlength\tabcolsep{2.5pt}
  \begin{tabular}{|c||c||c|c|c||c|c|c||c|c|c||c|c|c|}
    \hline
    & \multirow{2}{*}{Method}
    & \multicolumn{3}{c||}{$\sigma$=75} & \multicolumn{3}{c||}{$\sigma$=50} & \multicolumn{3}{c||}{$\sigma$=35} & \multicolumn{3}{c|}{$\sigma$=25}\\\cline{3-14}
    & & PSNR & SSIM & 25$\%$-ile & PSNR & SSIM & 25$\%$-ile & PSNR & SSIM & 25$\%$-ile & PSNR & SSIM & 25$\%$-ile\\ \hline\hline

    \multirow{4}{*}{Urban-100}
    & \bmd & 25.97 & 0.784 & 24.09 & 27.94 & 0.843 & 25.96 & 29.27 & 0.875 & 27.07 & 31.38 & 0.912 & 29.20 \\\cline{2-14}
    & \ircnn & - & - & - & 27.69 & 0.842 & 25.63 & 29.50 & 0.881 & 27.39 & 31.20 & 0.911 & 29.06 \\\cline{2-14}
    & \ffdnet & 26.05 & 0.793 & 23.98 & 28.05 & 0.850 & 25.93 & 29.78 & 0.887 & 27.61 & 31.40 & 0.914 & 29.17 \\\cline{2-14}
    \rowcolor{lightgray!40}& Ours-Blind & - & - & - & 28.57 & 0.859 & 26.47 & 30.21 & 0.893 & 28.04 & 31.70 & 0.917 & 29.49 \\\cline{2-14}
    \rowcolor{lightgray}& Ours & \textbf{26.68} & \textbf{0.811} & \textbf{24.64} & \textbf{28.62} & \textbf{0.862} & \textbf{26.52} & \textbf{30.26} & \textbf{0.895} & \textbf{28.09} & \textbf{31.81} & \textbf{0.919} & \textbf{29.59}  \\ \hline\hline

    \multirow{6}{*}{Kodak-24}
     & \bmd & 26.82 & 0.714 & 25.07 & 28.45 & 0.775 & 26.51 & 29.90 & 0.821 & 27.77 & 31.67 & 0.868 & 29.51 \\\cline{2-14}
     & \ircnn & - & - & - & 28.81 & 0.792 & 26.76 & 30.43 & 0.838 & 28.32 & 32.03 & 0.878 & 29.91 \\\cline{2-14}
    & \dncnn & 25.04 & - & - & 28.85 & - & - & 30.46 & - & - & 32.03 & - & - \\\cline{2-14}
     & \ffdnet & 27.27 & 0.733 & 25.30 & 28.98 & 0.793 & 26.89 & 30.57 & 0.841 & 28.41 & 32.13 & 0.879 & 29.95 \\\cline{2-14}
     \rowcolor{lightgray!40}& Ours-Blind & - & - & - & 29.21 & 0.803 & 27.11 & 30.78 & 0.849 & 28.63 & 32.31 & \textbf{0.884} & 30.14 \\\cline{2-14}
     \rowcolor{lightgray}& Ours & \textbf{27.56} & \textbf{0.748} & \textbf{25.59} & \textbf{29.25} & \textbf{0.805} & \textbf{27.15} & \textbf{30.81} & \textbf{0.849} & \textbf{28.66} & \textbf{32.34} & \textbf{0.884} & \textbf{30.19}  \\ \hline\hline

    \multirow{8}{*}{CBSD-68}
    &\bmd & 25.75 & 0.698 & 23.60 & 27.38 & 0.767 & 25.07 & 28.89 & 0.821 & 26.46 & 30.71 & 0.872 & 28.25\\\cline{2-14}
    & *CBM3D-Net~\cite{bm3dnet} & - & - & - & 27.48 & - & - & - & - & - & 30.91 & - & - \\\cline{2-14}
    & *CNL-Net~\cite{nlcnn} & - & - & - & 27.64 & - & - & - & - & - & 30.96 & - & - \\\cline{2-14}
    &\ircnn & - & - & - & 27.86 & 0.792 & 25.54 & 29.50 & 0.844 & 27.14 & 31.16 & 0.886 & 28.81\\\cline{2-14}
    & \dncnn & 24.47 & - & - & 27.92 & - & - & 29.58 & - & - & 31.23 & - & - \\\cline{2-14}
    &\ffdnet & 26.24 & 0.723 & 23.92 & 27.96 & 0.792 & 25.56 & 29.58 & 0.844 & 27.14 & 31.21 & 0.886 & 28.78 \\\cline{2-14}
    \rowcolor{lightgray!40}& Ours-Blind & - & - & - & 28.03 & 0.797 & 25.65 & 29.62 & 0.848 & 27.21 & 31.22 & \textbf{0.888} & 28.82 \\\cline{2-14}
    \rowcolor{lightgray}& Ours & \textbf{26.39} & \textbf{0.734} & \textbf{24.08} & \textbf{28.06} & \textbf{0.799} & \textbf{25.69} & \textbf{29.64} & \textbf{0.849} & \textbf{27.24} & \textbf{31.24} & \textbf{0.888} & \textbf{28.85} \\ \hline\hline

    \multirow{6}{*}{McMaster}
    &\bmd & 26.80 & 0.735 & 24.78 & 28.52 & 0.794 & 26.41 & 29.92 & 0.833 & 27.73 & 31.66 & 0.874 & 29.49 \\\cline{2-14}
    &\ircnn & - & - & - & 28.91 & 0.807 & 26.78 & 30.59 & 0.851 & 28.48 & 32.18 & 0.885 & 30.11 \\\cline{2-14}
    & \dncnn & 25.10 & - & - & 28.61 & - & - & 30.14 & - & - & 31.51 & - & - \\\cline{2-14}
    & \ffdnet & 27.33 & 0.760 & 25.19 & 29.18 & 0.816 & 26.99 & 30.81 & 0.857 & 28.62 & \textbf{32.35} & 0.889 & \textbf{30.20} \\\cline{2-14}
    \rowcolor{lightgray!40}& Ours-Blind & - & - & - & 29.31 & 0.824 & 27.14 & 30.85 & 0.861 & 28.69 & 32.31 & \textbf{0.890} & 30.14 \\\cline{2-14}
    \rowcolor{lightgray}& Ours & \textbf{27.59} & \textbf{0.775} & \textbf{25.47} & \textbf{29.35} & \textbf{0.826} & \textbf{27.16} & \textbf{30.90} & \textbf{0.863} &\textbf{ 28.70} & 32.33 & \textbf{0.890} & 30.17 \\ \hline
  \end{tabular}
  \footnotesize{*Other methods that are based on internal image statistics. See supplementary for comparisons of these to denoising with only our matching network.}
  \vspace{0.5em}
  \caption{Denoising Performance at Various Noise Levels on Different Datasets. We report performance in terms of Average PSNR (dB) and SSIM. To gauge robustness, we also report the $25^{th}$\%-ile worst-case PSNR (dB), computed across $8\times 8$ patches across each dataset. Ours-Blind refers to results from a common model of our method that is trained for a range of noise levels $\sigma\in[0,55]$ (and does not have knowledge of the specific noise level of its input).}
  \label{tab:allcolor}
\end{table*}

\section{Experiments}
\label{sec:exp}

\subsection{Preliminaries}

We train our algorithm on a set of 1600 color images from the Waterloo exploration dataset~\cite{waterloo}, and 168 images from the BSD-300~\cite{BSDS} train set, using the remaining 32 images for validation and parameter setting. We train our network using noisy observations generated by adding Gaussian noise to clean images in the training set. Unless otherwise specified, we construct the candidate set $\mathcal{N}_i$ of patches by considering all the overlapping patches in a $31\times 31$ search window around patch $i$.

We use Adam~\cite{adam} to train both the matching and regression networks, with an initial learning rate of $10^{-3}$. We pre-train the matching network for a 100k iterations based on the modified loss \eqref{eq:modloss}, with batches of 16 images and pairing all non-overlapping patches with randomly shuffled counterparts. This leads to a large number of ordered matching pairs for pre-training in each batch. We then continue training the matching network with the true loss in \eqref{eq:loss}, in this case forming a batch with 256 unique reference patches from various images, and computing matching scores for each with respect to all $31^2$ candidates. We train with this loss till saturation, with two learning rate drops of $10^{0.5}$. Once the matching network is trained, we store a set of noisy and denoised version of our training set, and use these to train the regression network (with the same training schedule, but without pre-training). Our code and trained models will be made available on publication.

\subsection{Denoising with Known Noise Level}
We evaluate our method for the task of color image denoising at five different noise levels, corresponding to additive white Gaussian noise with standard deviations of $\sigma=$25, 35, 50, and 75 gray levels. We train a separate network model for each level, and report their performance in Table \ref{tab:allcolor} on four datasets: Urban-100~\cite{urban100}, Kodak-24~\cite{kodak}, CBSD-68~\cite{foe}, and McMaster~\cite{mcmaster}.  For comparison, we also show results from a number of other state-of-the-art color denoising methods~\cite{ffdnet,ircnn,dncnn,bm3d,bm3dnet,nlcnn}. We evaluate performance in terms of the standard PSNR and SSIM~\cite{ssim} metrics, and to measure robustness, report worst-case errors as the $25^{th}$\%-ile among PSNR values of all individual $8\times 8$ patches in all images in each dataset. We find that our results are consistently more accurate across all datasets, with significant improvements over state-of-the-art methods at higher noise levels (with an improvement of 0.63 dB at noise-level 75 on the Urban-100 images). Moreover, not only are our denoised estimates more accurate on average in terms of PSNR and SSIM, our worst-case performance is also better---highlighting the robustness of our approach.

\begin{figure}[!t]
  \centering
  \includegraphics[width=0.95\columnwidth]{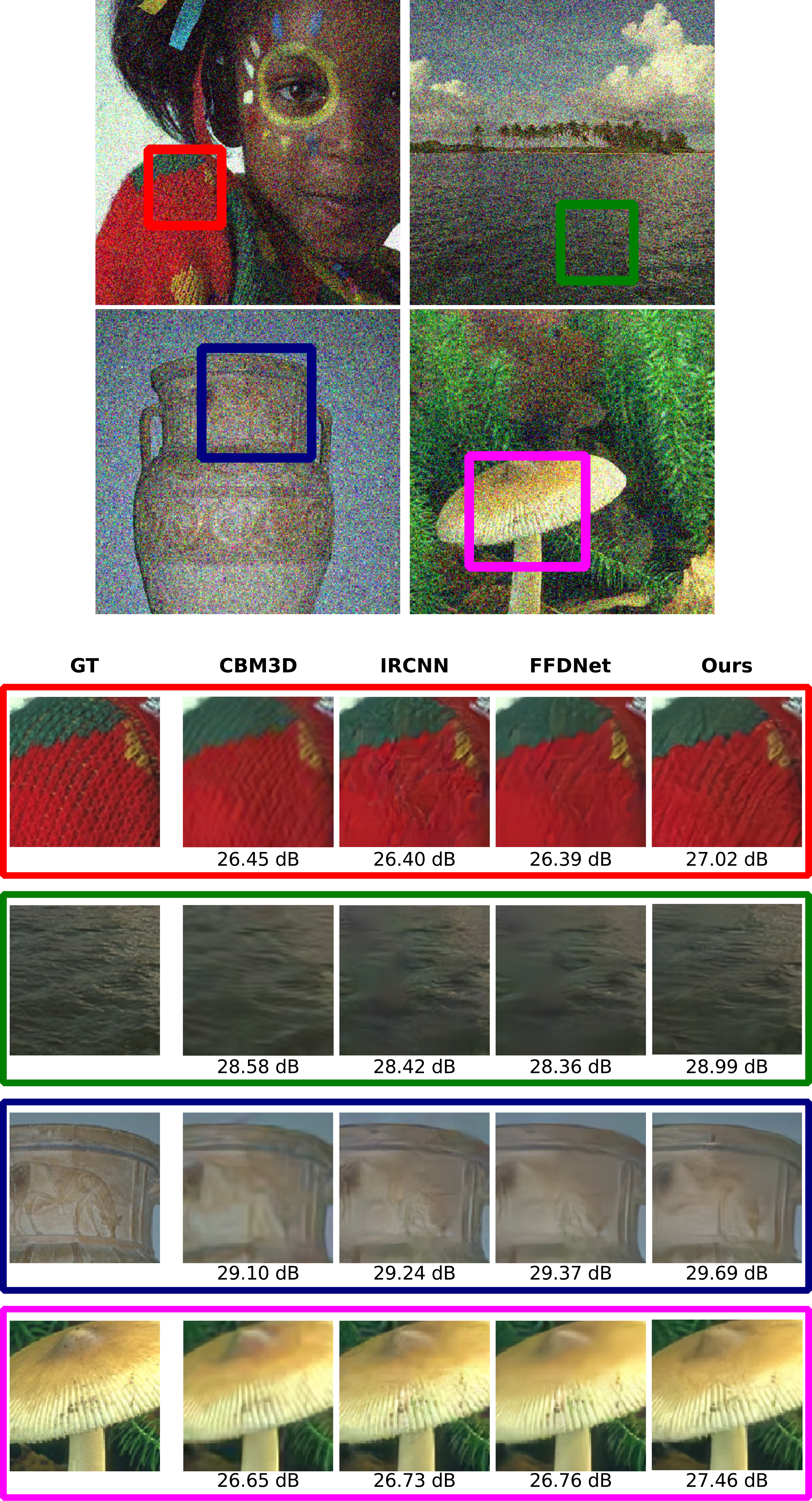}
  \caption{Example Crops of Denoised Images ($\sigma=50$). Compared to state-of-the-art denoising algorithms (IRCNN~\cite{ircnn}, DnCNN~\cite{dncnn}, and FFDNet~\cite{ffdnet}), we see that our overall method is often able to recover texture and detail with higher fidelity, by exploiting similar patterns in the input image itself.}
  \label{fig:res}
\end{figure}

We include examples of denoised images in Fig.~\ref{fig:res} for a qualitative evaluation, and see that denoised results from our method often contain  better reconstructions of texture and image detail than state-of-the-art denoising methods. In general, we find that our method has an advantage when a scene contains many repeating textures as expected, and also when it contains unique patterns---that are rare in training data and which regression-based methods are thus unable to reliably estimate. For images with limited repeating patterns, the burden of denoising then falls more to our second regression network, which then is able to still achieve results of acceptable quality at the level of standard regression-based methods.

\subsection{Blind Denoising}

Next, we consider the task of blind denoising, when the level of Gaussian noise in an observed image is unknown. For this, we follow the approach of \cite{dncnn} in training a common model for a range of noise levels $\sigma\in [0,55]$, by adding Gaussian noise with $\sigma$ chosen randomly for each image during training. Note that unlike for FFDNet~\cite{ffdnet}, the noise level for a specific input image is \emph{not} provided to our model. Table~\ref{tab:allcolor} also includes an evaluation of this version of our method (as Ours-Blind). We find that its performance is only slightly lower than that of our noise-specific networks, and still better in almost all cases than that of state-of-the-art methods that are aware of the level of noise in their inputs. This represents an attractive and practically useful variant of our method---which does not require maintaining multiple models for each noise level, and can be applied even when the noise level is unknown.

\subsection{Training with Limited Data}

\begin{figure}[!t]
  \centering
  \includegraphics[width=\columnwidth]{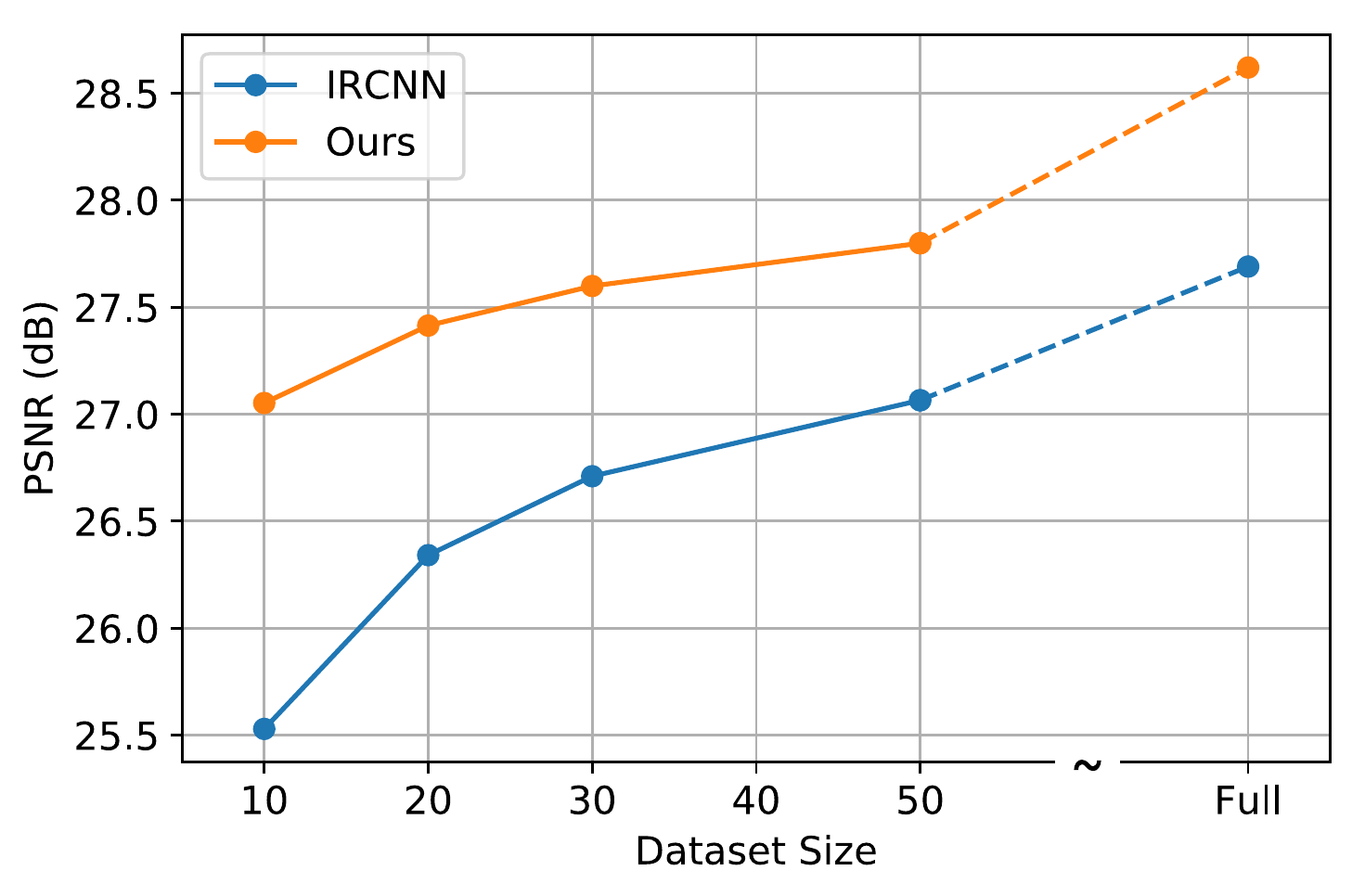}
  \caption{Effect of training set size. We report average PSNR on Urban-100~\cite{urban100} for denoising at $\sigma=50$ with our method and IRCNN~\cite{ircnn}, when both are trained with a limited number of training images (``Full'' represents using the entire training set for our method, and the official model for IRCNN). While our estimates are always more accurate than those from IRCNN, the gap is especially higher when the number of training images is small.}
  \label{fig:tsize}
\end{figure}

For many forms of image data and measurement models (\eg, medical images), a large amount of training data is difficult to acquire. Here, our method presents an advantage because of its focus on leveraging common patterns and textures in the input image itself, rather than those it has observed previously in a training set. We demonstrate this advantage by comparing our method to IRCNN~\cite{ircnn} in Fig.~\ref{fig:tsize}, when both methods are trained with only a few  training images (selected from our complete training set).

We assume a fixed known noise level of $\sigma=50$, and train versions of both models with different training set sizes---ranging from 10 to 50 images. Since overfitting is an issue with so little data, we track the performance of both methods on a validation set of 32 images through training, and choose the version with highest validation accuracy. We do not drop the learning rate for either method in this setting. Figure~\ref{fig:tsize} shows the average PSNR for both methods on the Urban-100 dataset~\cite{urban100}, for different training set sizes. While our method outperforms IRCNN~\cite{ircnn} in all cases, the performance gap is notably larger for smaller training sets---at 1.5 dB when training with only 10 images.

\subsection{Matching Network Analysis}
\label{sec:match}

Our matching network is a key component of our denoising algorithm. We end by analyzing its performance when applied to neighborhoods of different sizes, and the role that pre-training plays in convergence to a good solution. We also visualize the matching scores it generates, and how these differ across different groups of sub-bands. Furthermore, the supplementary includes evaluations of denoising using just the matching network (i.e., without the second regression network) on all datasets.

\pars{Effect of Window Size and Pre-training.} In Table~\ref{tab:ablat}, we characterize the trade-off between quality and computational cost when choosing different search window sizes over which to match and average patches. For different window sizes, we report average PSNR (over our validation set) for our initial match-averaged estimates when training with a known noise level of $\sigma=25$. We also report the corresponding running time required to compute matching scores and perform the averaging for different window sizes---for a $256\times 256$ input image on an NVIDIA 1080Ti GPU. Note that computing the initial estimates takes a majority of the time in our denoising method---the following regression step takes only an additional 0.01 seconds, and is independent of window size.

As expected, running time goes up roughly linearly with the number of candidate matches (i.e., as square of the search window size), but we find that the drop in PSNR is a modest $0.06$ dB when going down to a $23\times 23$ window. Table~\ref{tab:ablat} also demonstrates the importance of pre-training, and reports performance (again, of our initial estimates) achieved by a network that is initialized with random weights instead of with pre-training. We find that this leads to a PSNR drop of about 0.1 dB, highlighting that pre-training is important for convergence to a good model.

\begin{figure*}[!t]
  \centering
  \includegraphics[width=0.84\textwidth]{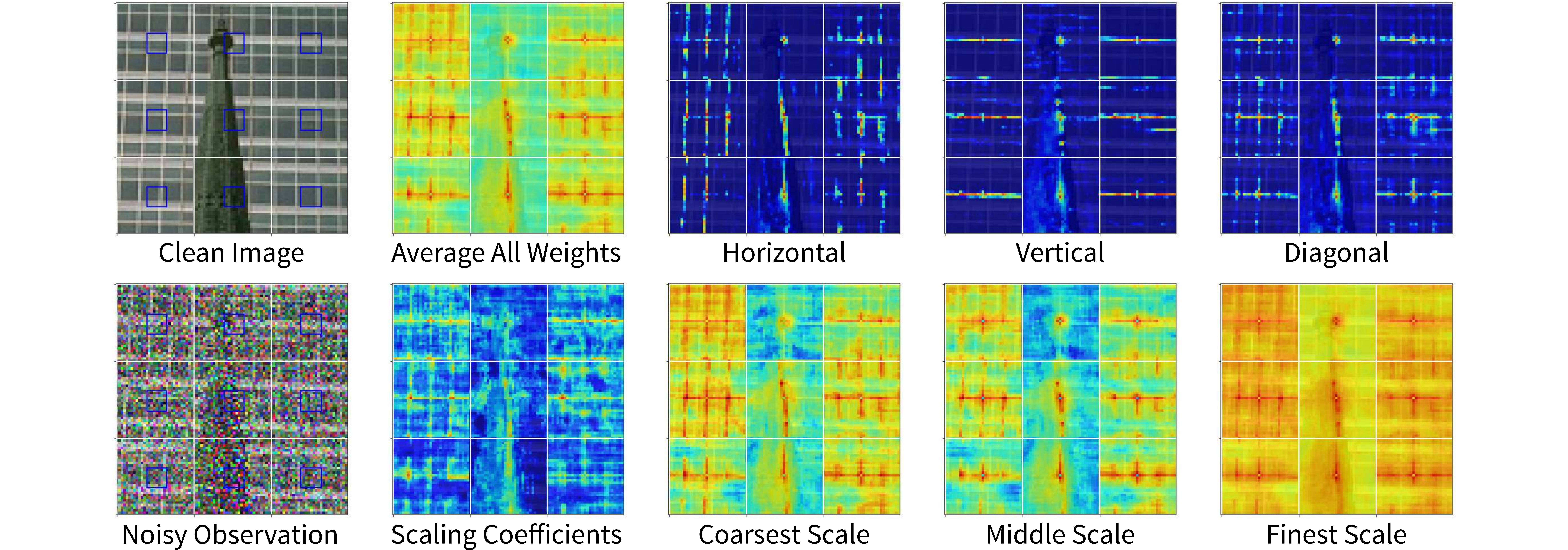}\\~\\
  \includegraphics[width=0.84\textwidth]{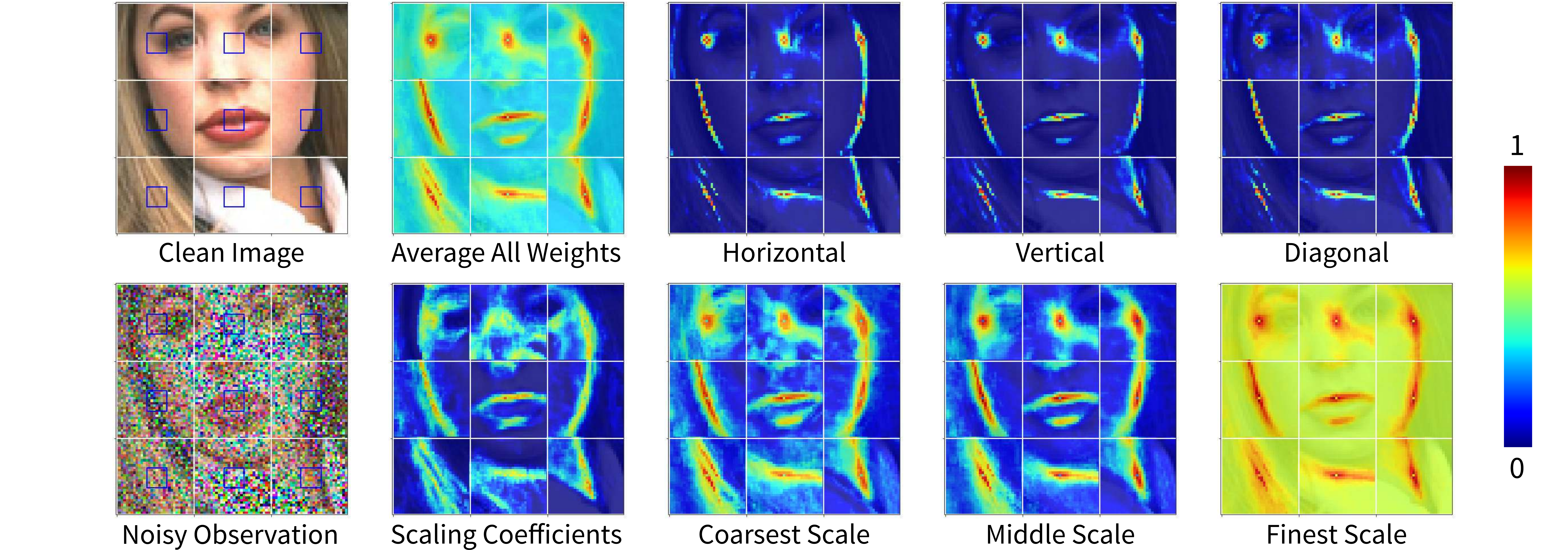}\\~\\
  \includegraphics[width=0.84\textwidth]{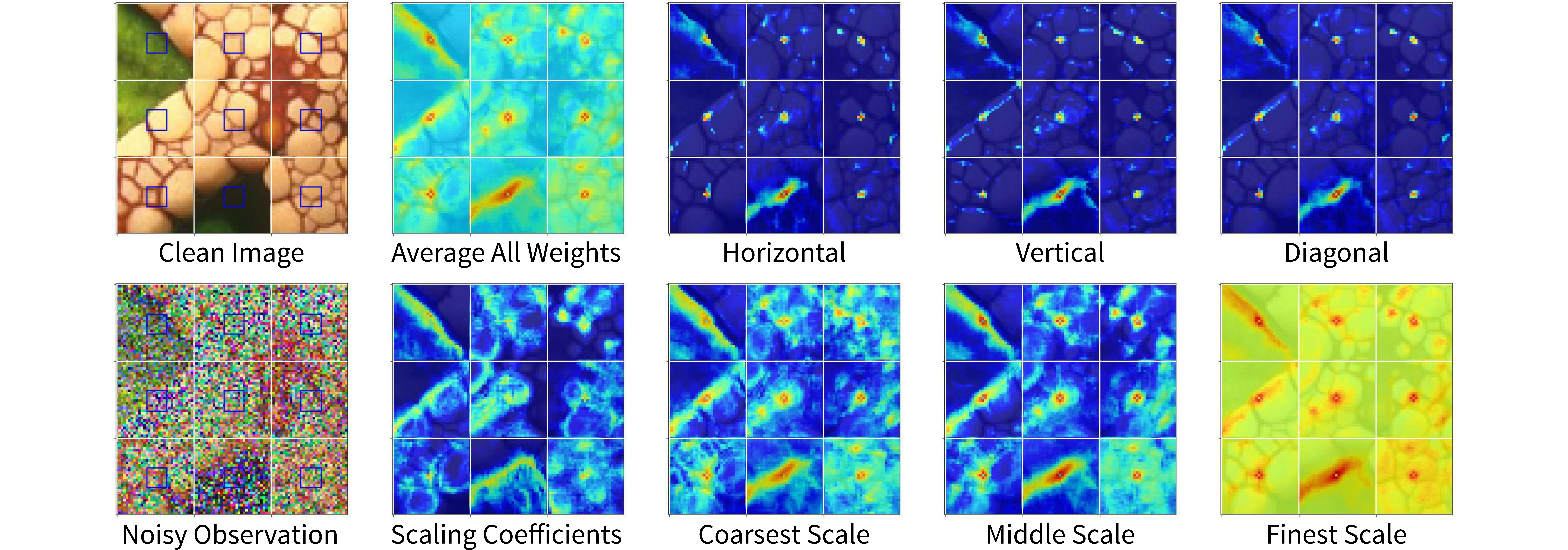}
  \caption{Visualization of Matching Score Distributions in Different Sub-bands. We show reference patches (indicated by blue squares) along with their local search windows from various images of the training set (9 windows per image), and visualize the matching scores predicted by our network. We show the predicted weights averaged across all sub-bands, as well as specific to different scales (averaging over color and orientation at each scale), and orientations (averaging over color and scale).}
  \label{fig:viz}
\end{figure*}

\begin{table}[!t]\small
  \centering
  \setlength\tabcolsep{4pt}
  \begin{tabular}{|l||c|c|c||c|}\hline
    Window Size & 15 & 23 & 31 & 31 (No Pre-training)\\\hline\hline
    PSNR (dB)  & 31.31 & 31.40 & 31.46 & 31.35\\\hline
    Run Time & 1.07s & 2.47s & 4.42s & 4.42s\\\hline
  \end{tabular}
  \vspace{0.5em}
  \caption{Window Size and Pre-Training Ablation. We report average PSNR (db) of the initial match-averaged estimates from our method on a validation set for $\sigma=25$. Run-times are for $256\times 256$ images on a 1080Ti GPU.}
  \label{tab:ablat}
\end{table}
\pars{Visualizing Matching Scores.} For a number of reference patches cropped from different training images, and their corresponding search windows,  we visualize the matching scores predicted by our network in Fig.~\ref{fig:viz}. We show the average matching score across all sub-bands, as well as average weights corresponding to combinations of sets at the same wavelet scale (averaging over color channels and orientation), and at the same orientation (averaging over scale and color). We see that the matching network produces very different averaging weights for different sub-bands.

We find that that the weights tend to be generally higher at the finest scale (indicating more averaging), and lowest for the scaling coefficients. This is likely because the highest-frequencies are close to zero in most patches, and thus to each other. For lower-frequencies and DC values, the network selects only those patches that are close to the reference patch (in the clean image). For different orientations, the high matches are sometimes concentrated at different locations for the same reference, especially when there are strong edges and repeating textures. Thus, free from the contraint of matching patches as a whole with a single score, our network finds different sets of matches for different sub-bands in order to achieve optimal denoising.

\section{Conclusion}
\label{sec:conc}

In this work, we presented a denoising method that employed a neural network to identify and exploit recurring patterns in an observed noisy image. Our network provided a fine-grained characterization of similarity, in terms of separate scores for different corresponding sub-band components, and thus enabled the recovery of high-quality denoised estimates. We also showed that our network is especially useful in regimes where training data is scarce, being able to achieve relatively higher performance from training on a small number of examples than standard regression-based methods. A natural direction of future work lies in exploring applications of our approach, of characterizing sub-band level self-similarity, to other image-like signals such as depth maps and motion-fields.

\noindent\textbf{Acknowledgments.} This work was supported by the National Science Foundation under award no. IIS-1820693.

{\small

}

\clearpage\onecolumn\appendix
\pagenumbering{roman}

\section{Separate Evaluation of Matching Network}

While our main evaluation considers the performance of our overall method, below we separately evaluate the performance of just our matching network and compare it to other ``internal statistics''-based methods. Our matching network is trained with the objective of maximizing denoising quality when using its outputs as weights for averaging patches. Therefore, as evaluation, we include average PSNR and SSIM values on all datasets of the \emph{initial} estimates of our method: based on averaging using predicted matching scores (but without the second regression step). For comparison, we also include the results of the other internal statistics-based methods from Table 1: CBM3D which is based on sum-of-squares distance (SSD) matching, and the neural network-based methods CBM3D-Net and CNL-Net.

We find that even our matching network by itself outperforms past self similarity-based methods (while our full method achieves state-of-the-art performance as demonstrated in the main paper).

{
\begin{center}
  \begin{tabular}{|c||c||c|c||c|c||c|c||c|c|}
    \hline
    & \multirow{2}{*}{Method}
    & \multicolumn{2}{c||}{$\sigma$=75} & \multicolumn{2}{c||}{$\sigma$=50} & \multicolumn{2}{c||}{$\sigma$=35} & \multicolumn{2}{c|}{$\sigma$=25}\\\cline{3-10}
    & & PSNR & SSIM & PSNR & SSIM & PSNR & SSIM & PSNR & SSIM\\ \hline\hline

    \multirow{2}{*}{Urban-100}
    & CBM3D & 25.97 & 0.784 & 27.94 & 0.843 & 29.27 & 0.875 & \textbf{31.38} & 0.912 \\\cline{2-10}
    & Ours: Match-average Only & \textbf{26.15} & \textbf{0.793} & \textbf{28.12} & \textbf{0.850} & \textbf{29.76} & \textbf{0.886} & {31.34} & \textbf{0.913}  \\ \hline\hline

    \multirow{2}{*}{Kodak-24}
     & CBM3D & 26.82 & 0.714 & 28.45 & 0.775 & 29.90 & 0.821 & 31.67 & 0.868 \\\cline{2-10}
     & Ours: Match-average Only & \textbf{27.27} & \textbf{0.735} & \textbf{28.98} & \textbf{0.796} & \textbf{30.53} & \textbf{0.843} & \textbf{32.06} & \textbf{0.880}  \\ \hline\hline

    \multirow{4}{*}{CBSD-68}
    &CBM3D & 25.75 & 0.698 & 27.38 & 0.767 & 28.89 & 0.821 & 30.71 & 0.872\\\cline{2-10}
    & CBM3D-Net & - & - & 27.48 & - & - & - & 30.91 & - \\\cline{2-10}
    & CNL-Net & - & - & 27.64 & - & - & - & 30.96 & - \\\cline{2-10}
    & Ours: Match-average Only & \textbf{26.15} & \textbf{0.723} & \textbf{27.83} & \textbf{0.791} & \textbf{29.40} & \textbf{0.843} & \textbf{31.00} & \textbf{0.884} \\ \hline\hline

    \multirow{2}{*}{McMaster}
    &CBM3D & 26.80 & 0.735 & 28.52 & 0.794 & 29.92 & 0.833 & 31.66 & 0.874 \\\cline{2-10}
    & Ours: Match-average Only & \textbf{27.18} & \textbf{0.757} & \textbf{28.92} & \textbf{0.812} & \textbf{30.39} & \textbf{0.850} & \textbf{31.81} & \textbf{0.882} \\ \hline
    \end{tabular}
  \end{center}}

\section{Additional Examples}
\subsection{Comparisons to FFDNet}
We begin by showing more visual results comparing our performance to the state-of-the-art method. Here, we include denoising estimates with both the ``blind'' and noise-specific versions of our model.

\noindent
\includegraphics[width=0.95\textwidth]{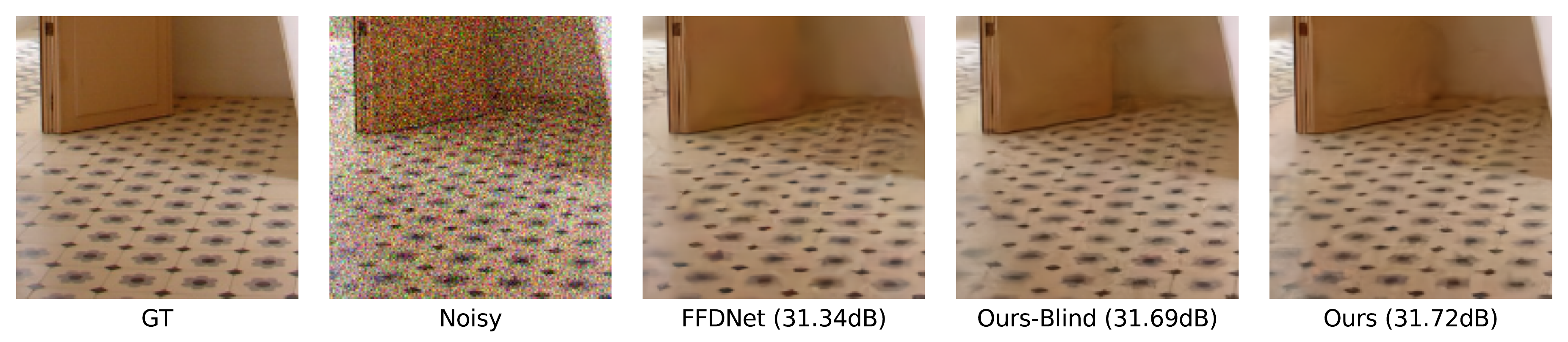}
\vspace{2em}

\includegraphics[width=0.95\textwidth]{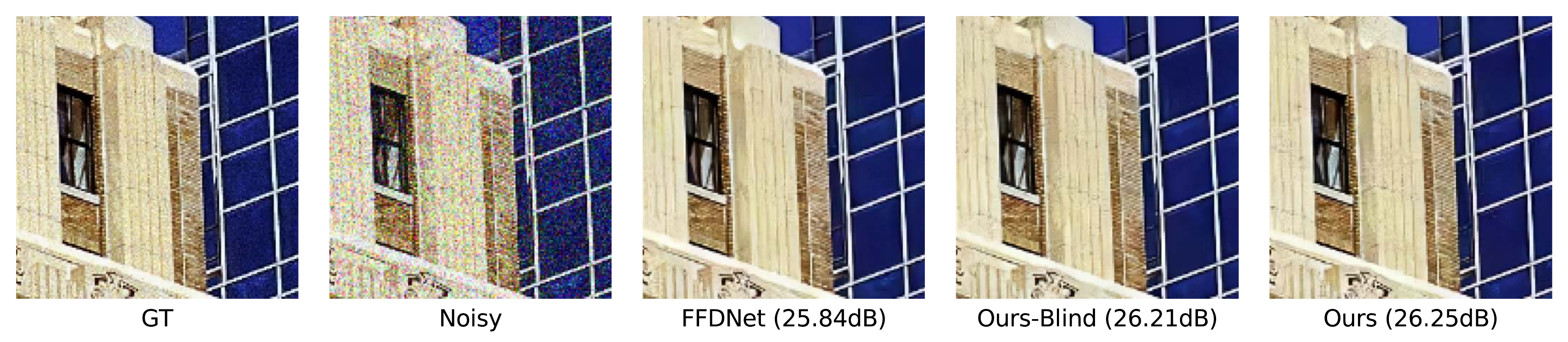}
\includegraphics[width=0.95\textwidth]{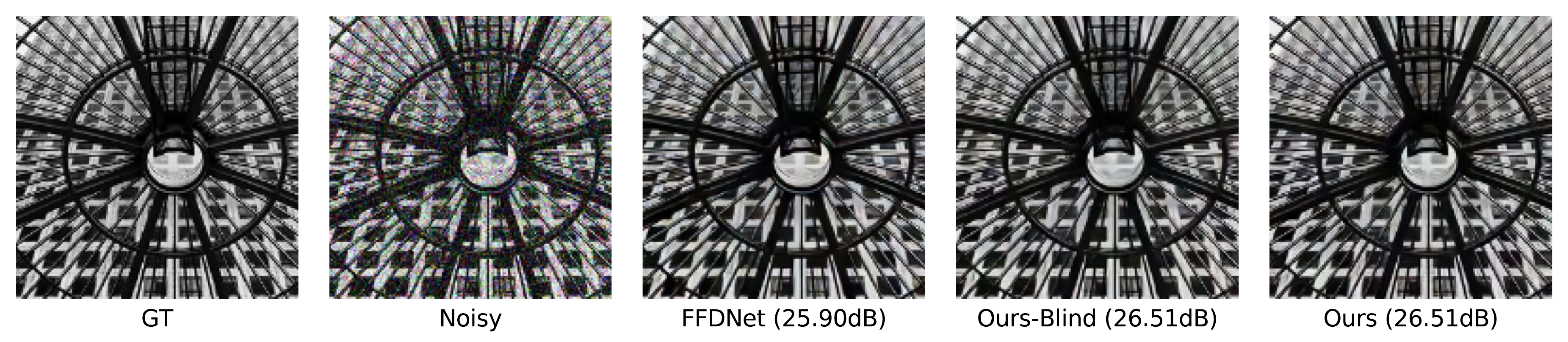}
\includegraphics[width=0.95\textwidth]{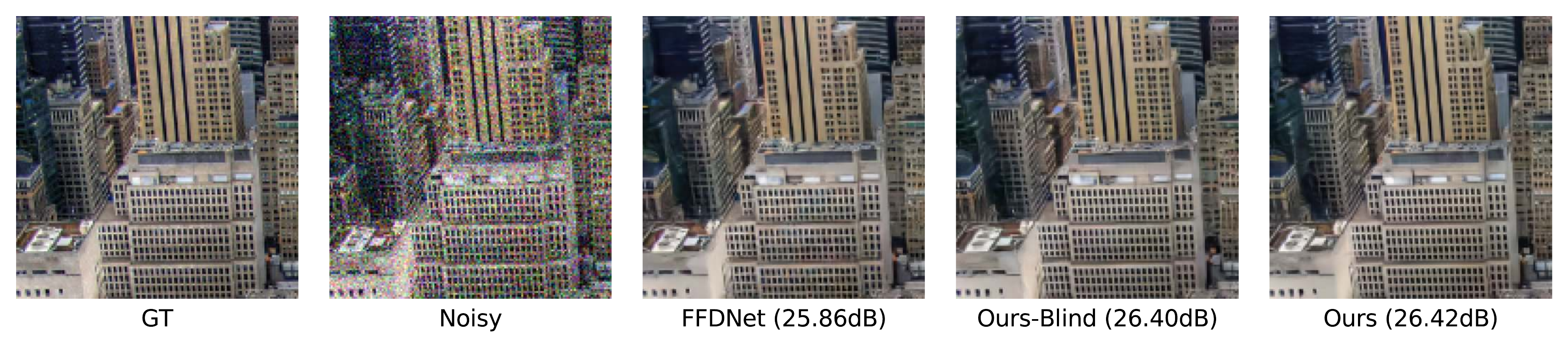}
\includegraphics[width=0.95\textwidth]{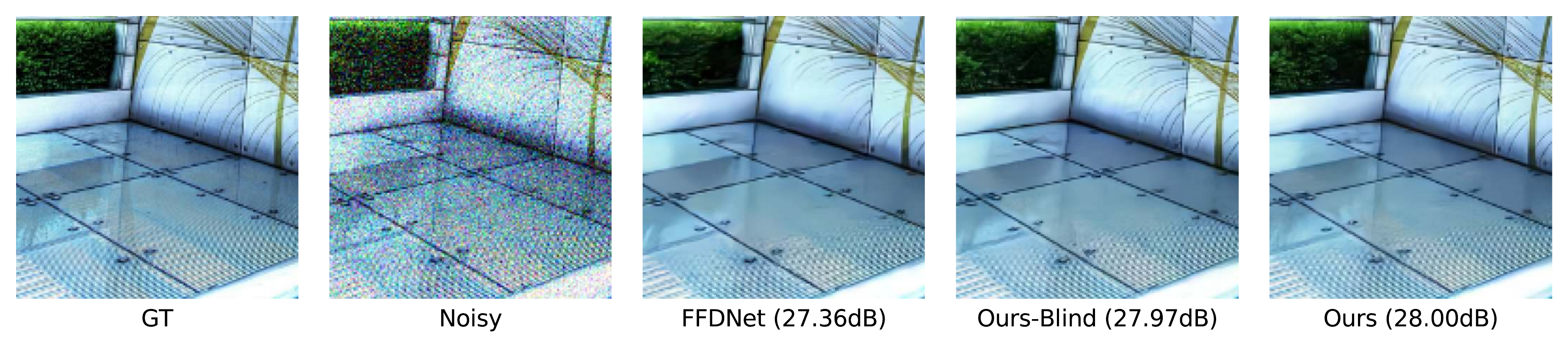}
\includegraphics[width=0.95\textwidth]{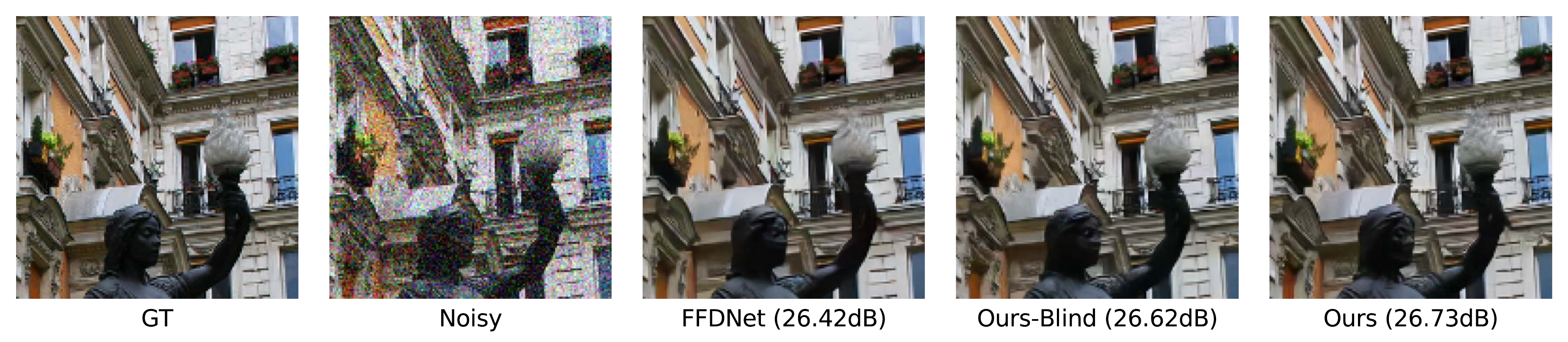}
\includegraphics[width=0.95\textwidth]{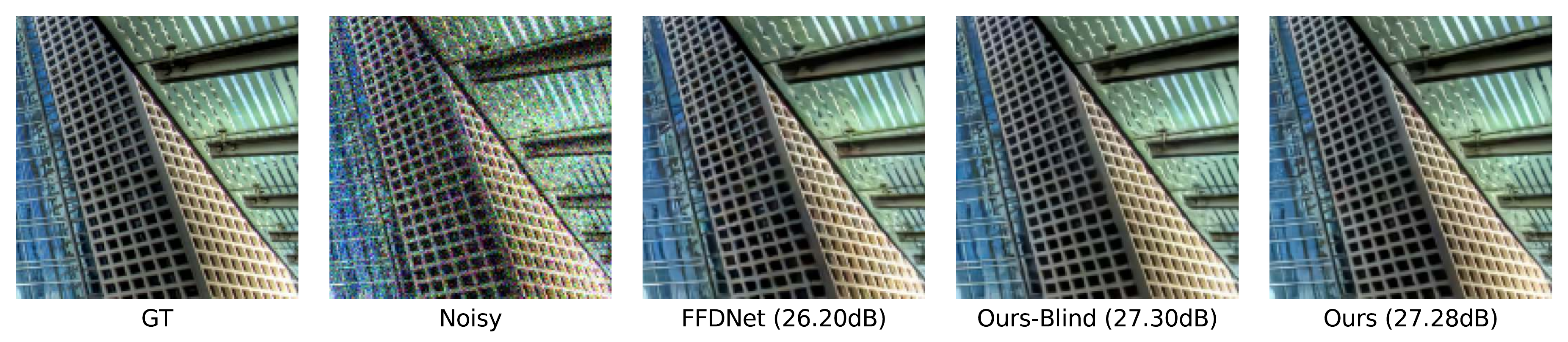}
\includegraphics[width=0.95\textwidth]{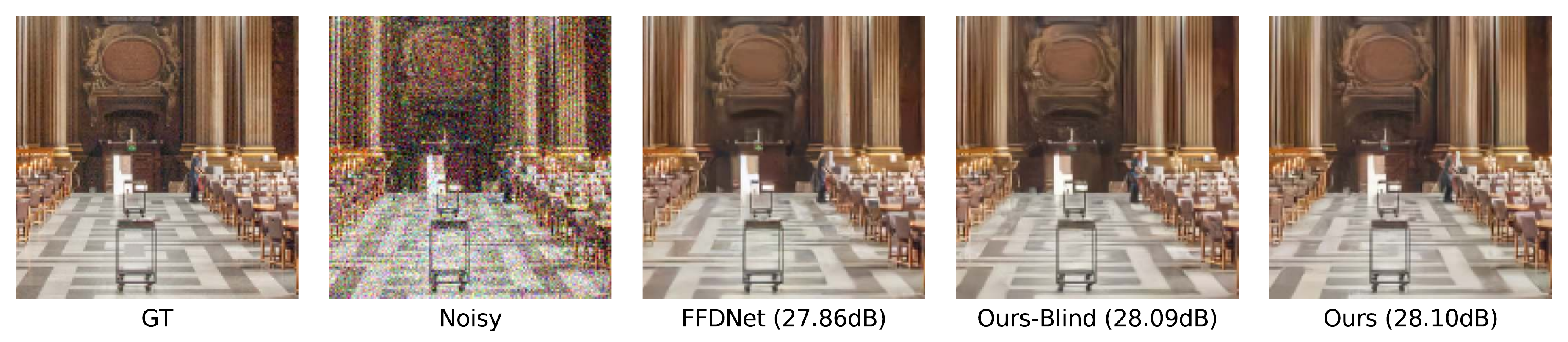}
\includegraphics[width=0.95\textwidth]{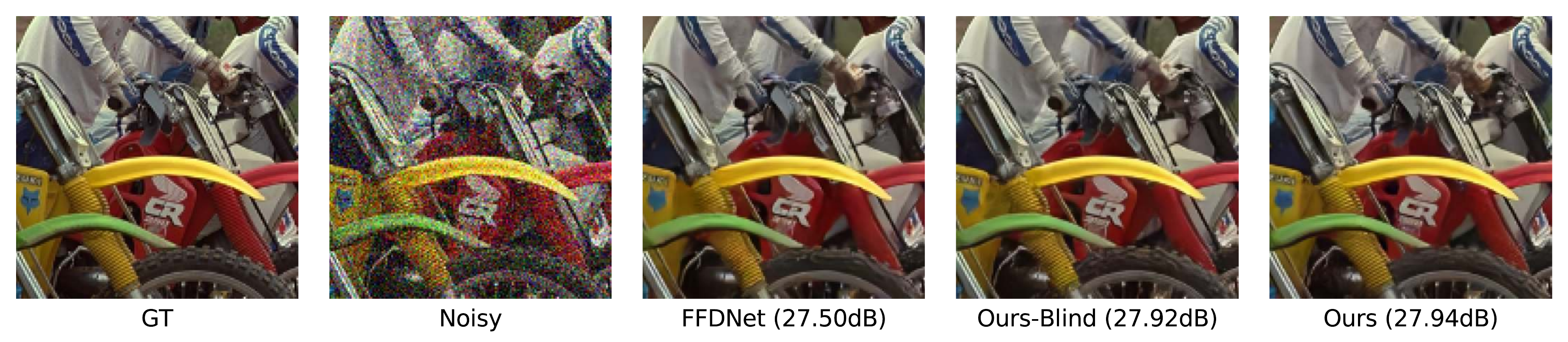}
\includegraphics[width=0.95\textwidth]{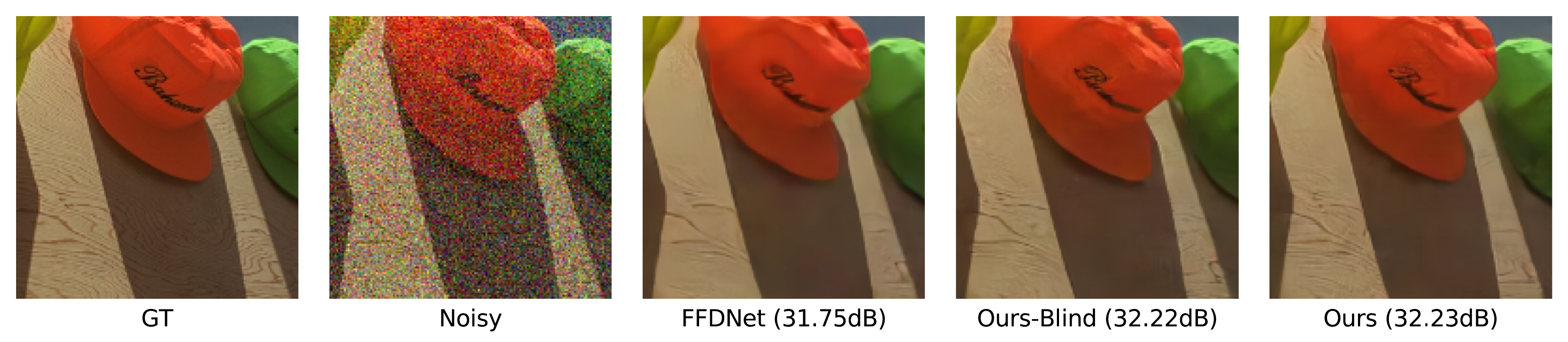}
\includegraphics[width=0.95\textwidth]{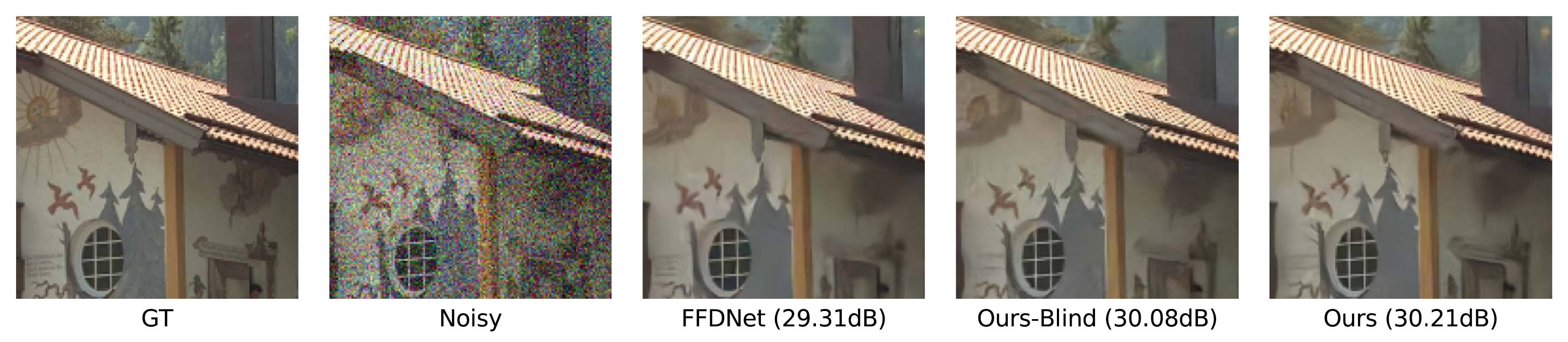}
\includegraphics[width=0.95\textwidth]{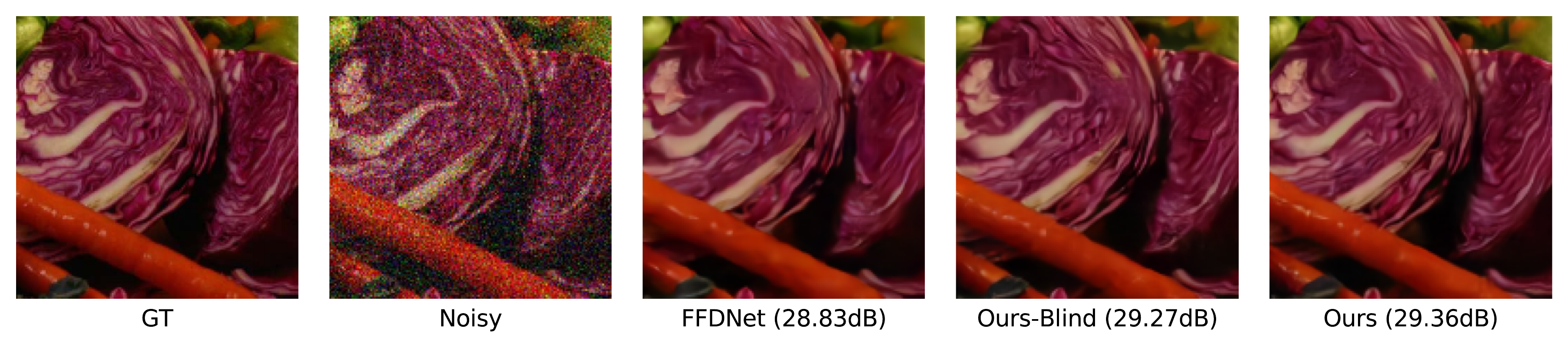}
\includegraphics[width=0.95\textwidth]{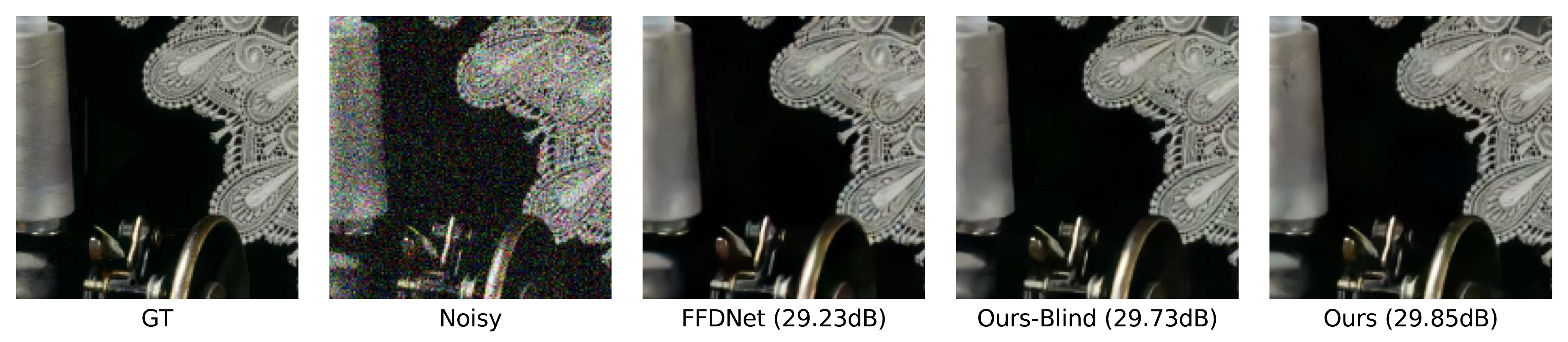}

\newpage
\subsection{Failure Cases}
Next, we show some of the examples of image regions where our denoised estimates have low accuracy.

\noindent
\includegraphics[width=0.95\textwidth]{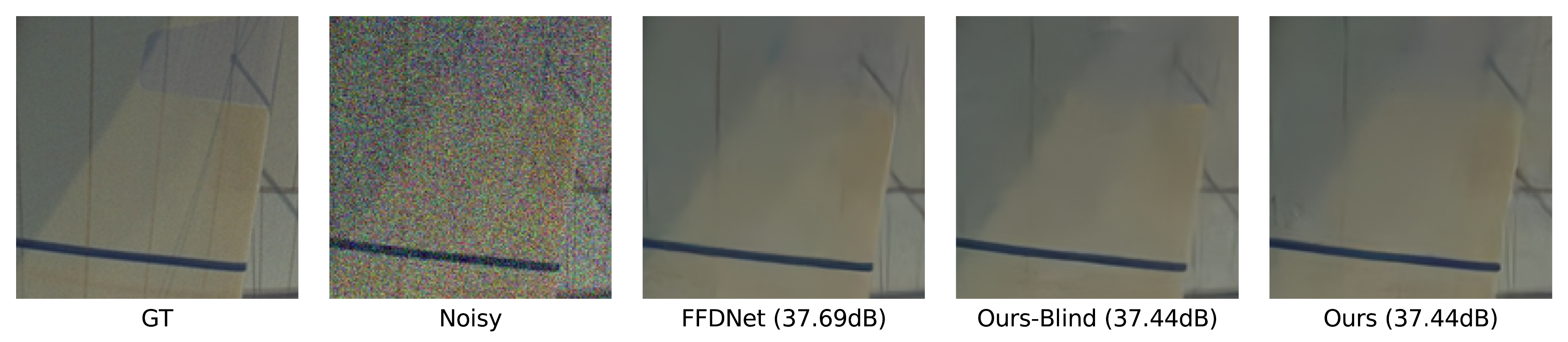}
\includegraphics[width=0.95\textwidth]{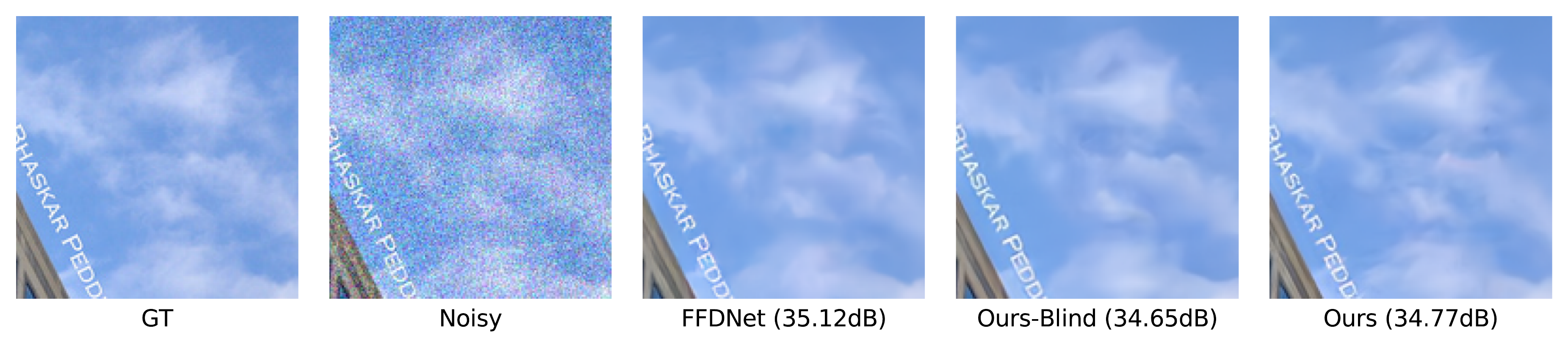}
\includegraphics[width=0.95\textwidth]{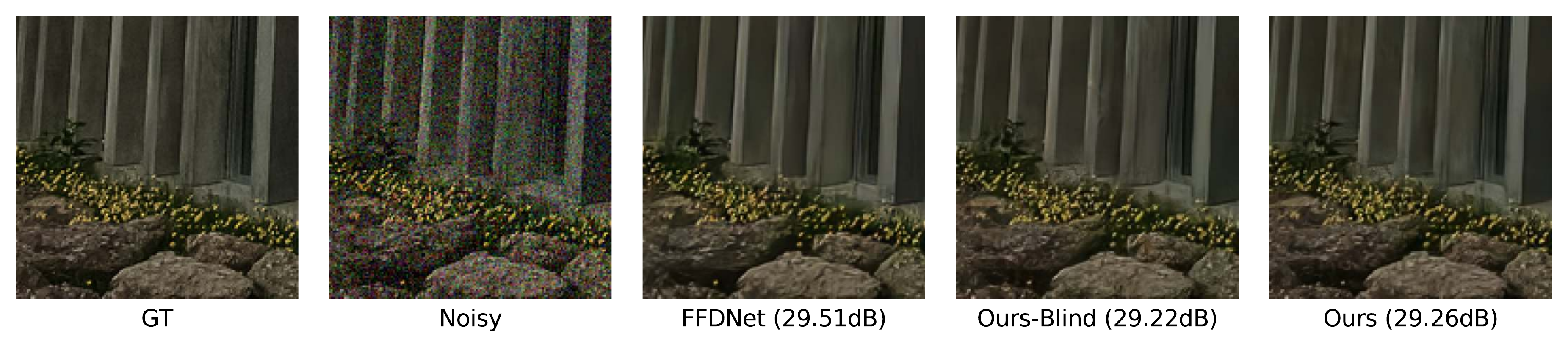}

\subsection{Initial vs Final Estimates}

Finally, we include examples of the intermediate output of our method---our initial estimates formed only by averaging based on scores from the matching network---and compare it to the final output after processing by the regression network. The match-average estimates are of reasonably high quality, and the regression network improves these results by varying amounts in different images (by removing subtle ``ringing-like'' artifacts).

\begin{center}
\includegraphics[width=0.76\textwidth]{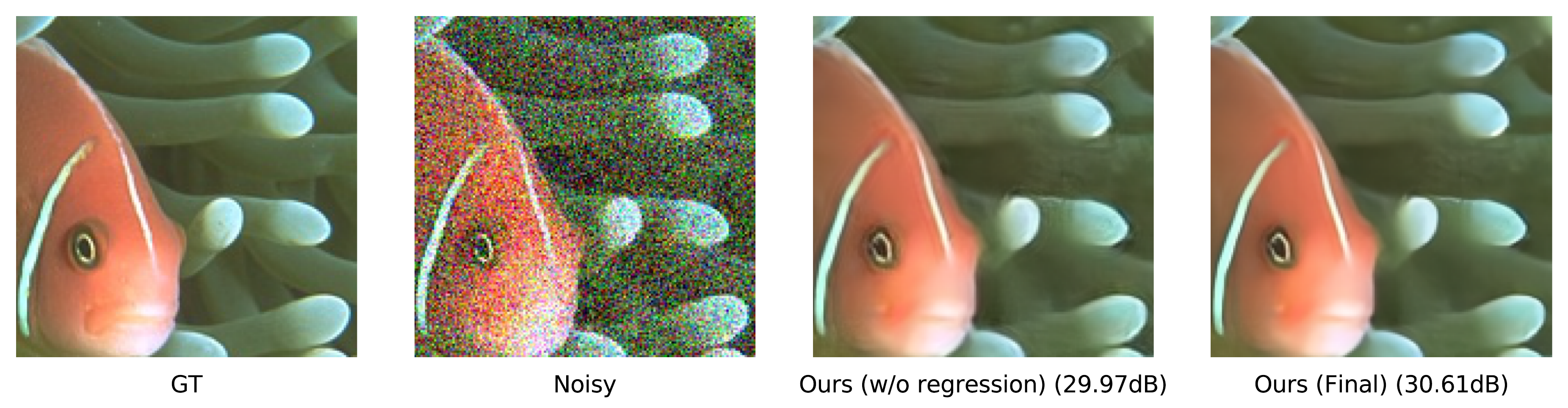}
\includegraphics[width=0.76\textwidth]{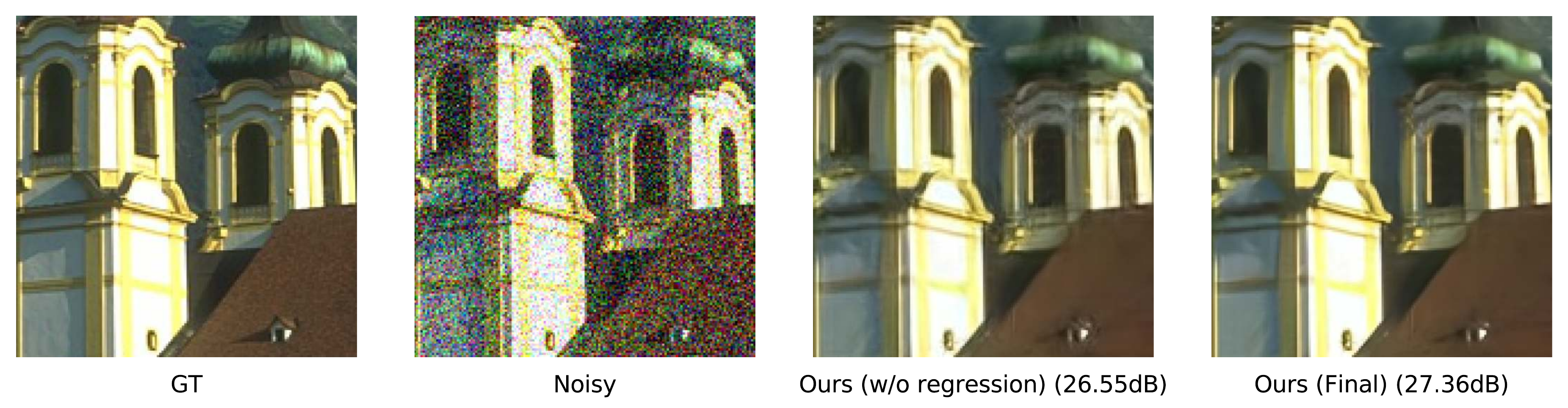}
\includegraphics[width=0.76\textwidth]{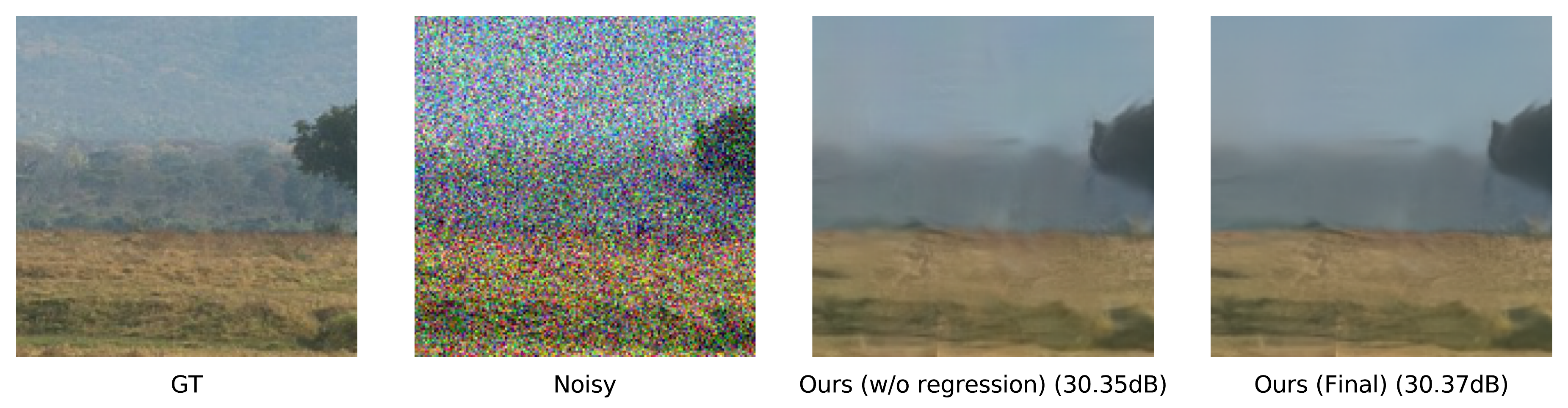}
\includegraphics[width=0.76\textwidth]{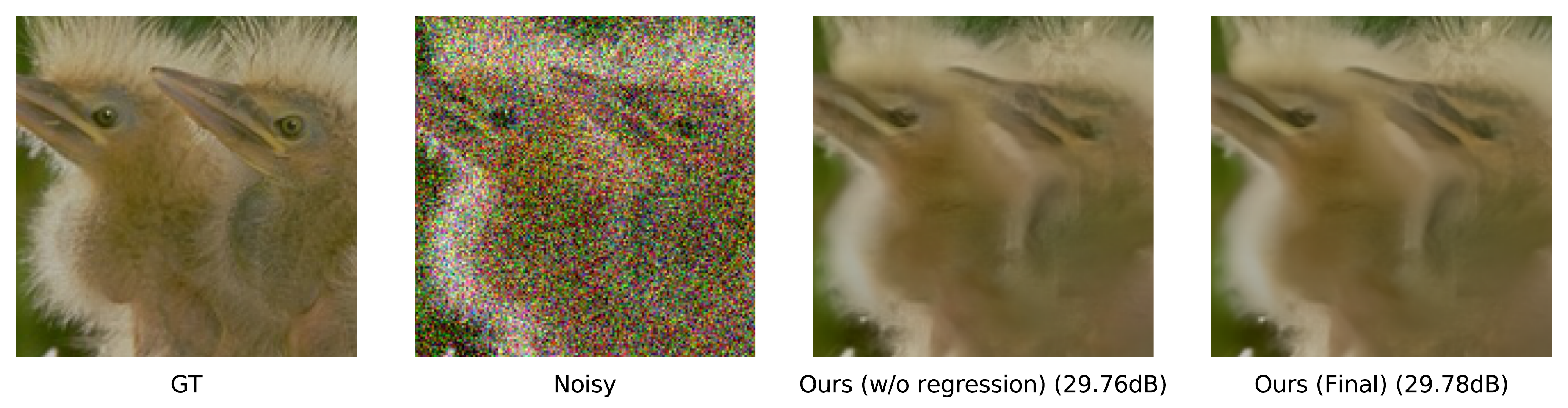}
\end{center}

\end{document}